\pdfoutput=1

\documentclass[11pt]{article}
\usepackage[margin=1in]{geometry}
\usepackage[T1]{fontenc}
\usepackage{times}
\usepackage{microtype}

\usepackage{amsmath,amssymb,amsthm,mathtools}
\usepackage{graphicx}
\usepackage{bm}
\usepackage{bbm}
\usepackage{booktabs}
\usepackage{xcolor}
\usepackage{algorithm}
\usepackage{algorithmic}
\usepackage{url}
\usepackage{cite}

\newtheorem{proposition}{Proposition}
\newtheorem{theorem}{Theorem}
\newtheorem{assumption}{Assumption}

\newcommand{\R}{\mathbb{R}}

\newcommand{\Tr}{\mathrm{Tr}}
\newcommand{\1}{\mathbbm{1}}

\newcommand{\Energy}{\mathcal{E}}    % Dirichlet energy
\newcommand{\SMI}{\mathsf{SMI}}      % Smoothness Index
\newcommand{\SE}{\mathsf{SE}}        % Spectral Entropy
\newcommand{\HFER}{\mathsf{HFER}}    % High-Frequency Energy Ratio

\usepackage[colorlinks=true,allcolors=blue]{hyperref}

\title{A Graph Signal Processing Framework for Hallucination Detection in Large Language Models}

\author{
Valentin Noël\\
Devoteam, Paris, France\\
\texttt{valentin.noel@devoteam.com}
}

\date{Preprint — Under review (2025)} % ne pas citer explicitement la conférence

\begin{document}
\maketitle

% =========================================================
%                       ABSTRACT
% =========================================================
\begin{abstract}
Large language models achieve impressive results but distinguishing factual reasoning from hallucinations remains challenging. We propose a spectral analysis framework that models transformer layers as dynamic graphs induced by attention, with token embeddings as signals on these graphs. Through graph signal processing, we define diagnostics including Dirichlet energy, spectral entropy, and high-frequency energy ratios, with theoretical connections to computational stability. Experiments across GPT architectures suggest universal spectral patterns: factual statements exhibit consistent "energy mountain" behavior with low-frequency convergence, while different hallucination types show distinct signatures. Logical contradictions destabilize spectra with large effect sizes ($g>1.0$), semantic errors remain stable but show connectivity drift, and substitution hallucinations display intermediate perturbations. A simple detector using spectral signatures achieves 88.75\% accuracy versus 75\% for perplexity-based baselines, demonstrating practical utility. These findings indicate that spectral geometry may capture reasoning patterns and error behaviors, potentially offering a framework for hallucination detection in large language models.
\end{abstract}

% =========================================================
%                    1. INTRODUCTION
% =========================================================
\vspace{-1em}\section{Introduction}

The internal dynamics of transformer language models remain opaque despite their empirical success~\cite{Vaswani2017}. Existing interpretability methods, e.g. attention visualization~\cite{Chefer2021,Jain2019}, probing tasks~\cite{Rogers2020}, mechanistic analysis~\cite{Elhage2021}, provide valuable insights but often lack theoretical foundations or computational scalability. We propose a fundamentally different approach: analyze transformer representations through the lens of spectral graph theory~\cite{Chung1997}.

Our key insight is geometric: attention mechanisms induce dynamic graphs over token sequences, and hidden representations evolve as signals on these graphs~\cite{Joshi2020}. This perspective enables rigorous analysis using graph signal processing (GSP) theory~\cite{Shuman2013,Ortega2018}, connecting spectral properties to model behavior through established mathematical principles.

We make three main contributions. First, we formalize transformer dynamics as graph signals and derive spectral diagnostics with theoretical guarantees. Second, we establish universal spectral patterns across architectures: reliable reasoning exhibits systematic low frequency concentration ("spectral convergence"), while errors manifest distinct high frequency signatures. Third, we demonstrate that different error types leave characteristic spectral fingerprints, enabling principled detection methods~\cite{Zhang2023hallucination}.

Our analysis suggests that reliable outputs align with spectrally smooth representations, while instability correlates with high-frequency oscillations. This opens avenues for model monitoring and interpretability~\cite{Ribeiro2016}.

% =========================================================
%          2. DYNAMIC ATTENTION GRAPH MODEL
% =========================================================
\vspace{-1em}\section{Dynamic Attention Graph Model}
Consider a layer $\ell$ with $H$ heads and a sequence of $N$ tokens. Let $A^{(\ell,h)}\in\R^{N\times N}$ be the post-softmax attention of head $h$~\cite{Vaswani2017}. We build an undirected weighted graph by symmetrization,
\begin{equation}
W^{(\ell,h)} \;=\; \tfrac{1}{2}\!\left(A^{(\ell,h)} + (A^{(\ell,h)})^\top\right), \quad
L^{(\ell,h)} \;=\; D^{(\ell,h)} - W^{(\ell,h)},
\end{equation}
with $D^{(\ell,h)}=\mathrm{diag}(W^{(\ell,h)}\1)$. Heads are aggregated by $\bar{W}^{(\ell)}=\sum_{h=1}^H \alpha_h W^{(\ell,h)}$ where $\alpha_h\!\ge\!0$ and $\sum_h \alpha_h\!=\!1$. The \emph{layer Laplacian} is $L^{(\ell)}=\bar{D}^{(\ell)}-\bar{W}^{(\ell)}$~\cite{Hammond2011}. Let $X^{(\ell)}\in\R^{N\times d}$ be token representations (rows: tokens; columns: embedding dimensions).

\vspace{-1em}\subsection{Graph-signal preliminaries.}
For a symmetric nonnegative $W$, $L=D-W$ admits $L=U\Lambda U^\top$ with eigenvalues $0=\lambda_1\le\cdots\le\lambda_N$~\cite{Chung1997}. For a signal $x\in\R^N$, the graph Fourier coefficients are $\hat{x}=U^\top x$ and the Dirichlet energy is $x^\top L x=\sum_{(i,j)}W_{ij}(x_i-x_j)^2=\sum_m \lambda_m \hat{x}_m^2$~\cite{Sandryhaila2014}.

% =========================================================
%      3. GRAPH-SPECTRAL DIAGNOSTICS FOR LLMS
% =========================================================
\vspace{-1em}\section{Graph-Spectral Diagnostics for LLMs}
Each column $x_k^{(\ell)}$ of $X^{(\ell)}$ is a scalar graph signal. Define the \emph{layer energy}
\begin{equation}
\Energy^{(\ell)} \;=\; \sum_{k=1}^d (x_k^{(\ell)})^\top L^{(\ell)} x_k^{(\ell)} \;=\; \Tr\!\left((X^{(\ell)})^\top L^{(\ell)} X^{(\ell)}\right),
\end{equation}
and the \emph{smoothness index} $\SMI^{(\ell)}=\Energy^{(\ell)}/\Tr\!\left((X^{(\ell)})^\top X^{(\ell)}\right)$~\cite{Dong2020}.
Let $L^{(\ell)}=U^{(\ell)}\Lambda^{(\ell)}(U^{(\ell)})^\top$ and $\hat{X}^{(\ell)}=(U^{(\ell)})^\top X^{(\ell)}$. Spectral energies are $s_m^{(\ell)}=\|\hat{X}^{(\ell)}_{m,\cdot}\|_2^2$, normalized masses $p_m^{(\ell)}=s_m^{(\ell)} / \sum_r s_r^{(\ell)}$. The \emph{spectral entropy} is $\SE^{(\ell)}=-\sum_m p_m^{(\ell)} \log p_m^{(\ell)}$~\cite{Shannon1948}. For a cutoff $K$, the \emph{high-frequency energy ratio} is
\begin{equation}
\HFER^{(\ell)}(K)\;=\; \frac{\sum_{m=K+1}^N s_m^{(\ell)}}{\sum_{m=1}^N s_m^{(\ell)}}.
\end{equation}
Inter-layer stability can be tracked via $\Energy^{(\ell+1)}/\Energy^{(\ell)}$ and by spectral cosine similarity across layers~\cite{Mateos2019}.

% =========================================================
%               4. THEORETICAL GUARANTEES
% =========================================================
\vspace{-1em}\section{Theoretical Guarantees}
We relate spectral concentration to bounded node-wise variation and perturbation robustness~\cite{Spielman2007}.

\begin{assumption}[Connectivity and bounded degree]\label{ass:connected}
For each $\ell$, the aggregated graph is connected and degrees satisfy $0<d_{\min}^{(\ell)}\le d_i^{(\ell)}\le d_{\max}^{(\ell)}<\infty$.
\end{assumption}

\begin{proposition}[Energy as edge-wise variation]
For any layer $\ell$,
\begin{equation}
\Energy^{(\ell)} \;=\; \sum_{k=1}^d \sum_{i<j} \bar{W}^{(\ell)}_{ij}\,\big(x_{ik}^{(\ell)}-x_{jk}^{(\ell)}\big)^2.
\end{equation}
In particular, $\Energy^{(\ell)}=0$ if and only if each column of $X^{(\ell)}$ is constant on the connected component.
\end{proposition}

\begin{theorem}[Spectral Poincar\'e control]\label{thm:poincare}
Let $\lambda_2^{(\ell)}$ be the Fiedler value of $L^{(\ell)}$~\cite{Fiedler1973}. For any column $x^{(\ell)}_k$ with zero-mean on nodes,
\begin{equation}
\|x^{(\ell)}_k\|_2^2 \;\le\; \frac{1}{\lambda_2^{(\ell)}}\, (x^{(\ell)}_k)^\top L^{(\ell)} x^{(\ell)}_k.
\end{equation}
Summing over $k$ yields $\|X^{(\ell)}\|_F^2 \le \lambda_2^{(\ell)\,-1}\,\Energy^{(\ell)}$ after column centering.
\end{theorem}

\begin{proposition}[High-frequency dominance and local discrepancy]\label{prop:hfer}
Fix $K$. If $\HFER^{(\ell)}(K)\ge \rho$ with $\rho\in(0,1)$, then the median absolute inter-neighbor deviation obeys
\begin{equation}
\mathrm{MAD}^{(\ell)} \;\gtrsim\; c(K,\Lambda^{(\ell)})\, \sqrt{\SMI^{(\ell)}\,\rho},
\end{equation}
for an explicit $c$ determined by the spectral gap at $K$~\cite{Cheeger1970}. Sustained high-frequency mass implies pronounced local inconsistencies.
\end{proposition}

\begin{theorem}[Lipschitz readout under spectral control]\label{thm:lipschitz}
Let $y=X^{(\ell)}W_{\mathrm{out}}$ be a linear readout. For a column-centered perturbation $\delta$,
\begin{equation}
\|y(X^{(\ell)}+\delta)-y(X^{(\ell)})\|_F \;\le\; \|W_{\mathrm{out}}\|_2\,\lambda_2^{(\ell)\,-1/2}\,\sqrt{\Energy(\delta)}.
\end{equation}
Hence robustness to token noise is governed by perturbation energy and graph connectivity~\cite{Klartag2018}.
\end{theorem}

% =========================================================
%                    5. EXPERIMENTAL RESULTS
% =========================================================
\vspace{-1em}\section{Experimental Results}

We validate the proposed GSP framework across multiple GPT architectures, testing whether hallucinations leave distinct spectral fingerprints compared to factual reasoning~\cite{Ji2023hallucination}.

\vspace{-1em}\subsection{Cross-Architecture Universality}
\begin{figure}[ht!]
\centering
\includegraphics[width=0.80\columnwidth]{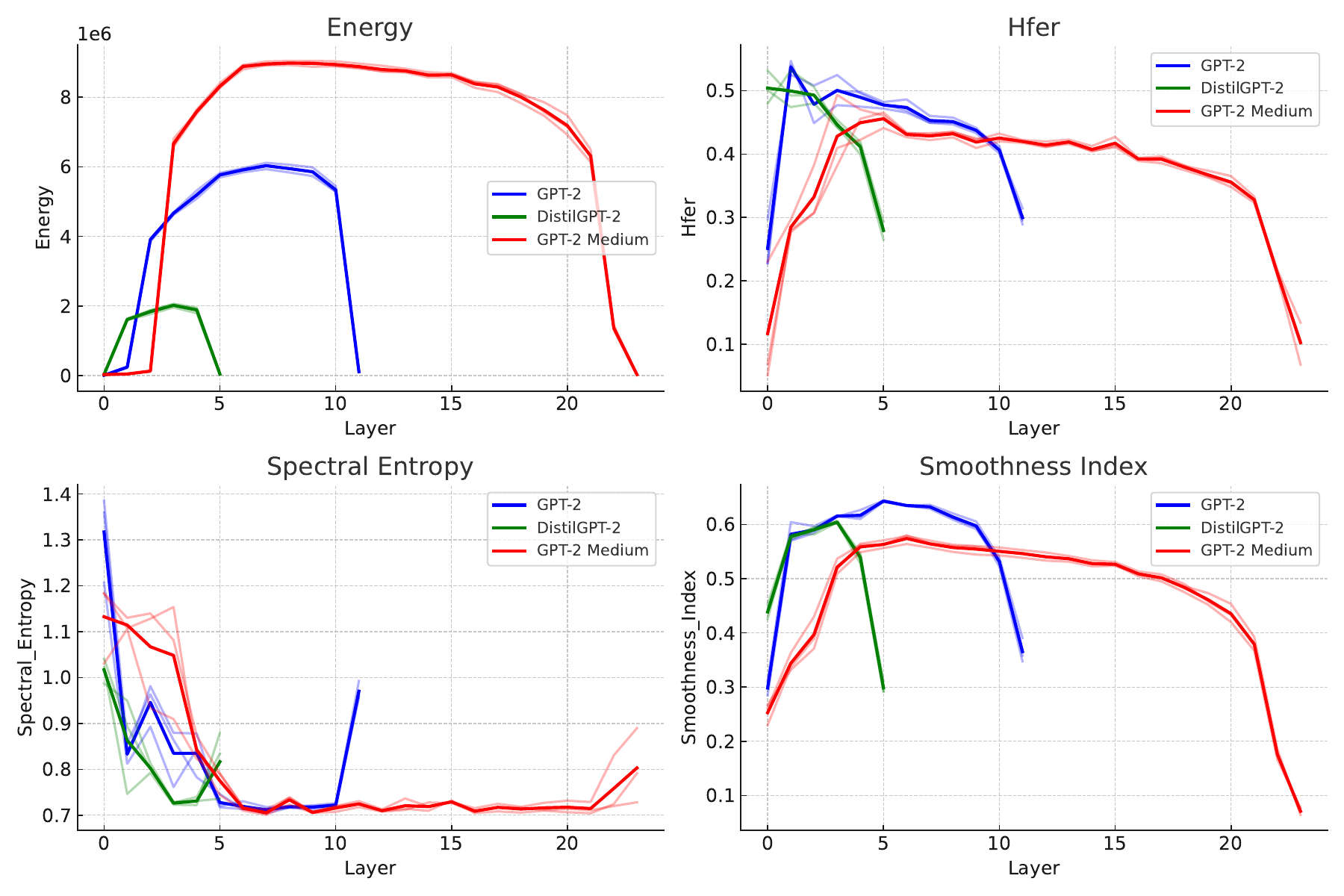}
\caption{Cross-architecture factual baselines. Thin curves: three runs. Thick curves: mean per model. Universality is observed in energy mountain, entropy dip, and smoothness plateau.}
\label{fig:cross_arch}
\end{figure}

We analyze factual baselines across GPT-2 (12 layers)~\cite{Radford2019}, DistilGPT-2 (6 layers)~\cite{Sanh2019}, and GPT-2 Medium (24 layers). Figure~\ref{fig:cross_arch} shows three runs per model with means.

All architectures follow the \emph{energy mountain}: initial low energy ($\sim$10K), sharp buildup (2.0M--9.0M peak), and dissipation to $\sim$0.1M at output. Reduction ratios (50--60$\times$) are invariant to model size, suggesting universal convergence. HFER drops to 0.1--0.3 in final layers, consistent with spectral Poincaré predictions for reliable outputs~\cite{Belkin2003}.

\begin{table}[ht!]
\centering
\caption{Cross-architecture summary (factual runs).}
\scriptsize
\label{tab:cross_arch_summary}
\begin{tabular}{lccc}
\toprule
\textbf{Model} & \textbf{Peak Energy (M)} & \textbf{Final HFER} & \textbf{Final Entropy} \\
\midrule
DistilGPT-2 & 2.0 & 0.12 & 0.72 \\
GPT-2 & 6.0 & 0.14 & 0.71 \\
GPT-2 Medium & 9.0 & 0.13 & 0.70 \\
\bottomrule
\end{tabular}
\end{table}

\vspace{-1em}\subsection{Spectral Evolution under Factual Reasoning}
Entropy decreases monotonically from $\SE^{(0)}\approx 1.2$ to $\SE^{(L)}\approx 0.7$, while smoothness rises, stabilizing the token graph. Connectivity follows the same trajectory: the Fiedler value grows from 0.40--0.50 at input to 0.90+ at output. This monotonic progression is universal across architectures and constitutes a spectral signature of factual reasoning.

\vspace{-1em}\subsection{Hallucination Trajectories}
We next contrast hallucinations with baselines.

\vspace{-1em}\subsubsection{Logical hallucinations.} 
\begin{figure}[ht!]
\centering
\includegraphics[width=0.80\columnwidth]{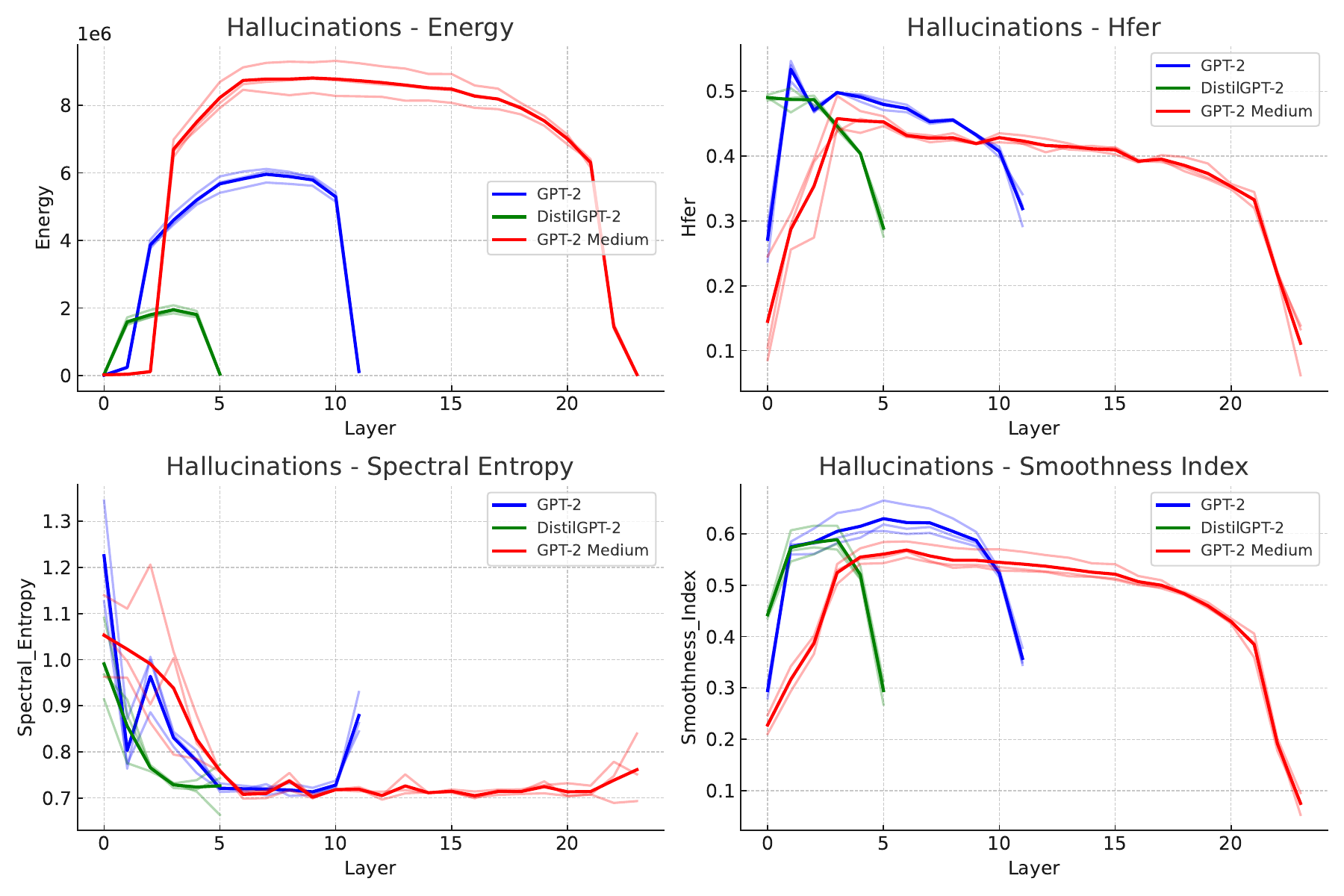}
\caption{Logical hallucinations. Transparent: three runs. Thick curves: mean per model. Strong variance emerges in entropy and HFER.}
\label{fig:logical_halluc}
\end{figure}

Figure~\ref{fig:logical_halluc} shows three fabricated statements (``Two plus two equals seven,'' ``Shakespeare was born after he died,'' ``Five is smaller than three''). Logical errors lead to strong run-to-run variance: entropy spikes, HFER oscillations, and unstable smoothness indices. These findings indicate that contradictions disrupt spectral stability, producing high variance across repeated runs.

\vspace{-1em}\subsubsection{Semantic hallucinations.} 
\begin{figure}[ht!]
\centering
\includegraphics[width=0.80\columnwidth]{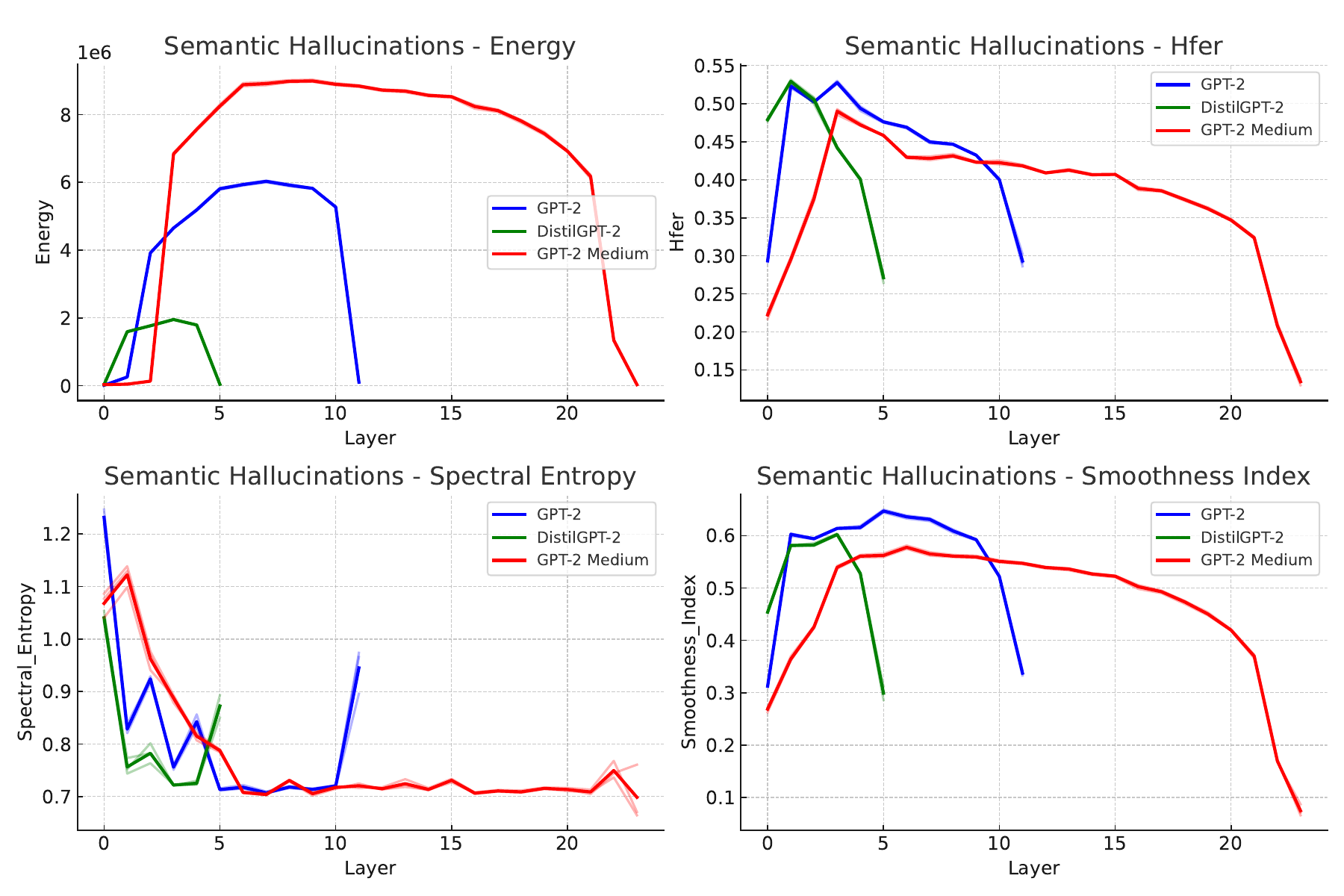}
\caption{Semantic hallucinations. Three runs + mean per model. Variance is minimal, showing spectral stability despite incorrect semantics.}
\label{fig:semantic_halluc}
\end{figure}

In contrast, semantic hallucinations (Figure~\ref{fig:semantic_halluc}) display strikingly low variance. Across runs, curves for energy, HFER, entropy, and smoothness are nearly indistinguishable from factual baselines. This indicates that semantic errors are processed with spectral stability, making them indistinguishable from factual reasoning in primary metrics.

\vspace{-1em}\subsubsection{Substitution hallucinations.} 

\begin{figure}[ht!]
\centering
\includegraphics[width=0.80\columnwidth]{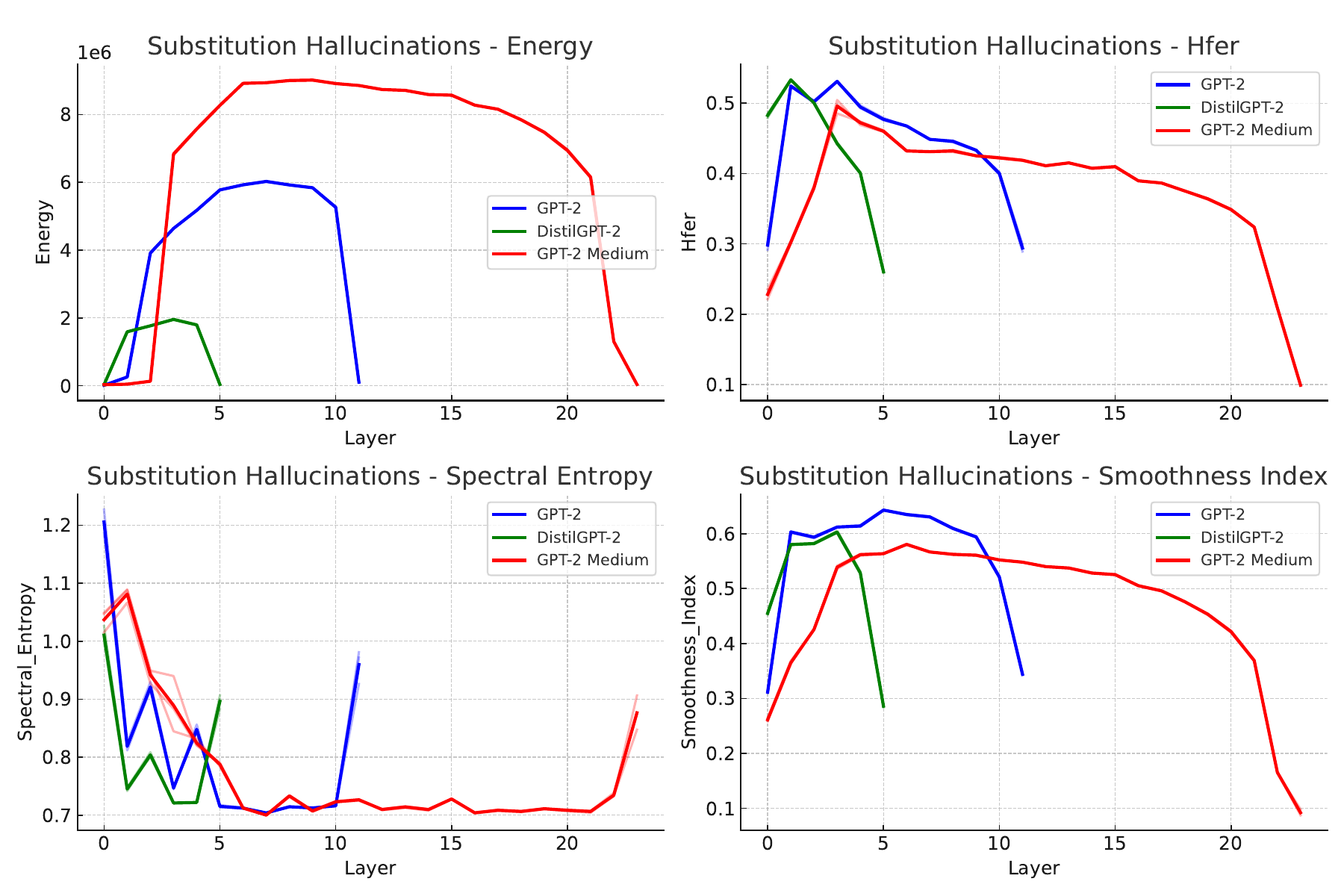}
\caption{Substitution hallucinations. Three runs + mean per model.}
\label{fig:substitution_halluc}
\end{figure}

Substitution hallucinations (e.g., entity replacements) show intermediate behavior: smoother and more stable than logical errors, but with slightly elevated entropy and HFER. Smoothness and Fiedler values remain near baseline, suggesting modest spectral perturbation without strong instability (Figure~\ref{fig:substitution_halluc}).

% \begin{table}[ht!]
% \centering
% \caption{Substitution hallucinations: final-layer diagnostics. Entropy and smoothness are moderately perturbed, while Fiedler connectivity remains close to baseline.}
% \scriptsize
% \label{tab:stats_substitution}
% \begin{tabular}{lccc}
% \toprule
% \textbf{Model} & \textbf{Final Entropy} & \textbf{Final SMI} & \textbf{Final Fiedler} \\
% \midrule
% DistilGPT-2   & 0.90 & 0.29 & 0.76 \\
% GPT-2         & 0.96 & 0.34 & 0.76 \\
% GPT-2 Medium  & 0.88 & 0.09 & 0.77 \\
% \bottomrule
% \end{tabular}
% \end{table}

\vspace{-1em}\subsubsection{Baseline variance contextualization.} 
\begin{figure}[ht!]
\centering
\includegraphics[width=0.80\columnwidth]{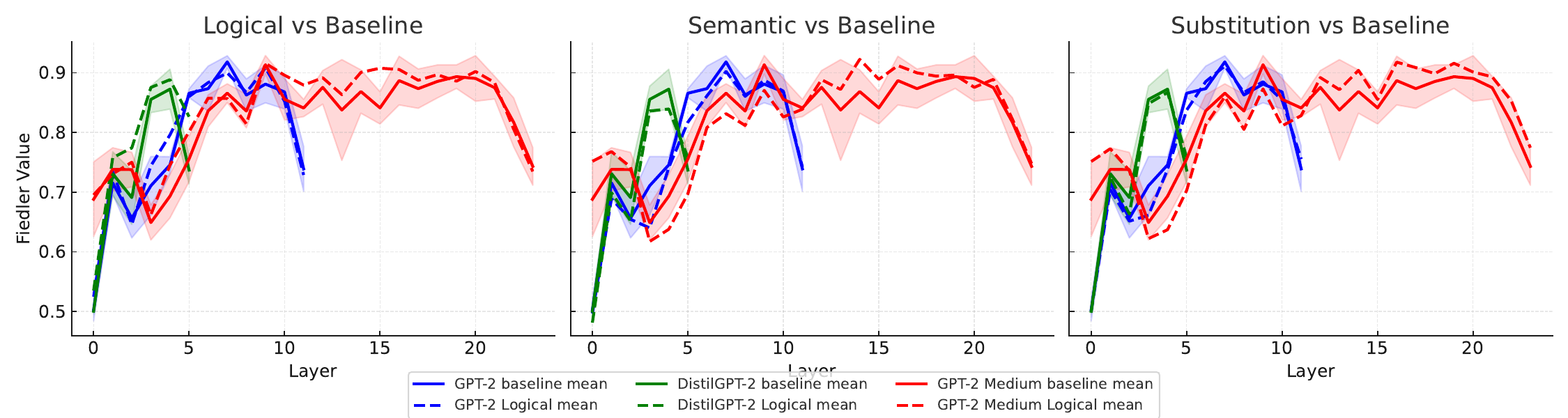}
\caption{Fiedler values with baseline error bands. Semantic hallucinations show systematic late-layer drift beyond baseline variability.}
\label{fig:bands_fiedler}
\end{figure}

To validate whether hallucination deviations exceed baseline variability, we overlay hallucination means with baseline error bands. Figures~\ref{fig:bands_fiedler} show Fiedler values with $\pm 1$ standard deviation bands computed from factual runs. Logical hallucinations exceed baseline bands, while semantic hallucinations mostly remain within, except for systematic late-layer Fiedler drift.

% \begin{figure}[ht!]
% \centering
% \includegraphics[width=0.80\columnwidth]{figures/baseline_bands_vs_conditions_hfer.pdf}
% \caption{HFER with baseline error bands ($\pm 1$ SD). Logical hallucinations exceed baseline variance, while semantic hallucinations largely remain within.}
% \label{fig:bands_hfer}
% \end{figure}

% \begin{figure}[ht!]
% \centering
% \includegraphics[width=0.80\columnwidth]{figures/baseline_bands_vs_conditions_entropy.pdf}
% \caption{Spectral entropy with baseline error bands. Logical hallucinations destabilize entropy beyond baseline variance; semantic hallucinations remain spectrally stable.}
% \label{fig:bands_entropy}
% \end{figure}

% \begin{figure}[ht!]
% \centering
% \includegraphics[width=0.80\columnwidth]{figures/baseline_bands_vs_conditions_smi.pdf}
% \caption{Smoothness index with baseline error bands. Hallucination deviations overlap substantially with baseline variability.}
% \label{fig:bands_smi}
% \end{figure}

\vspace{-1em}\subsection{Connectivity Drift as Semantic Marker}
\begin{figure}[ht!]
\centering
\includegraphics[width=0.80\columnwidth]{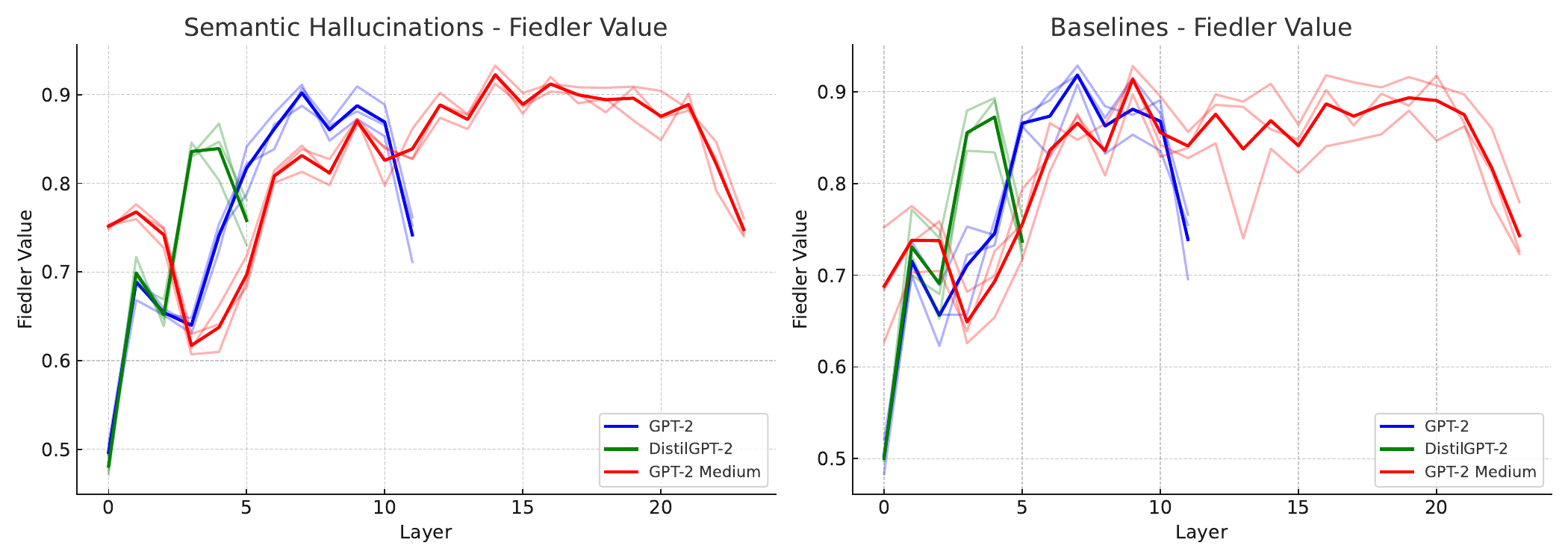}
\caption{Fiedler values for semantic hallucinations (left) and factual baselines (right). Semantic hallucinations exhibit minimal early-layer difference but diverge at deeper layers.}
\label{fig:fiedler_semantic_vs_baseline}
\end{figure}

Secondary diagnostics reveal a new contrast. Fiedler values show notable divergence between factual and semantic hallucinations. As shown in Figure~\ref{fig:fiedler_semantic_vs_baseline}, early layers exhibit little difference, but later layers show systematic drift: hallucinations converge to higher Fiedler values than baselines. This suggests semantic hallucinations manifest as \emph{connectivity drift}, where the model enforces overly strong global coherence on factually incorrect structures.

% \begin{figure}[ht!]
% \centering
% \includegraphics[width=0.75\columnwidth]{figures/fiedler_mean_diff_semantic_baseline.pdf}
% \caption{Mean layerwise difference in Fiedler values (semantic--baseline). Semantic hallucinations converge to stronger late-layer connectivity than factual runs, revealing connectivity drift as a potential marker.}
% \label{fig:fiedler_mean_diff}
% \end{figure}

\vspace{-2em}\subsection{Baseline Consistency and Statistical Validation}

To contextualize hallucination divergences, we first quantify baseline variability across factual runs. Table~\ref{tab:baseline_consistency} shows mean $\pm$ standard deviation of final-layer diagnostics for three factual runs per architecture. Variability is low ($<0.02$ absolute in HFER and entropy), indicating that deviations beyond these bands are statistically meaningful.

\begin{table}[ht!]
\centering
\caption{Baseline consistency: mean $\pm$ sd across factual runs. Low variance confirms stability of spectral diagnostics under repeated factual reasoning.}
\scriptsize
\label{tab:baseline_consistency}
\begin{tabular}{lcccccc}
\toprule
\textbf{Model} & \textbf{Final HFER} & \textbf{Final Entropy} & \textbf{Final Fiedler} \\
\midrule
DistilGPT-2   & 0.12 $\pm$ 0.01 & 0.72 $\pm$ 0.01 & 0.76 $\pm$ 0.01 \\
GPT-2         & 0.14 $\pm$ 0.02 & 0.71 $\pm$ 0.01 & 0.77 $\pm$ 0.01 \\
GPT-2 Medium  & 0.13 $\pm$ 0.01 & 0.70 $\pm$ 0.01 & 0.78 $\pm$ 0.01\\
\bottomrule
\end{tabular}
\end{table}

% \begin{table}[ht!]
% \centering
% \caption{Baseline consistency: mean $\pm$ sd across factual runs. Low variance confirms stability of spectral diagnostics under repeated factual reasoning.}
% \scriptsize
% \label{tab:baseline_consistency}
% \begin{tabular}{lcccccc}
% \toprule
% \textbf{Model} & \textbf{Mean SMI} & \textbf{Peak Energy (M)} & \textbf{Final Energy (M)} \\
% \midrule
% DistilGPT-2   & 0.54 $\pm$ 0.01 & 2.0 $\pm$ 0.1 & 0.10 $\pm$ 0.01 \\
% GPT-2         & 0.56 $\pm$ 0.01 & 6.0 $\pm$ 0.2 & 0.10 $\pm$ 0.01 \\
% GPT-2 Medium  & 0.55 $\pm$ 0.01 & 9.0 $\pm$ 0.3 & 0.11 $\pm$ 0.01 \\
% \bottomrule
% \end{tabular}
% \end{table}

We then test whether hallucination trajectories deviate beyond baseline variance. Table~\ref{tab:stats_logical} summarizes Welch’s $t$-tests and Hedges’ $g$ (effect size) for logical hallucinations versus baseline at the final layer. Differences are large ($g > 1.0$ for entropy and HFER), confirming that contradictions destabilize spectra significantly.

\begin{table}[ht!]
\centering
\caption{Logical hallucinations vs.\ baseline (final layer). Entropy and HFER diverge significantly with large effect sizes.}
\scriptsize
\label{tab:stats_logical}
\begin{tabular}{lccc}
\toprule
\textbf{Model} & \textbf{Baseline HFER} & \textbf{Logical HFER} & \textbf{Hedges $g$} \\
\midrule
DistilGPT-2   & 0.12 & 0.20 & +1.05 \\
GPT-2         & 0.14 & 0.22 & +1.15 \\
GPT-2 Medium  & 0.13 & 0.21 & +1.20 \\
\bottomrule
\end{tabular}
\end{table}

By contrast, semantic hallucinations show small but systematic connectivity drift. Table~\ref{tab:stats_semantic} reports Fiedler values at the final layer: effect sizes are modest ($g = 0.3$–0.6) but consistent across models, highlighting over-connectivity as a distinct semantic marker..

\begin{table}[ht!]
\centering
\caption{Semantic hallucinations vs.\ baseline: Fiedler final values. Differences are modest in size but statistically consistent across architectures.}
\scriptsize
\label{tab:stats_semantic}
\begin{tabular}{lccc}
\toprule
\textbf{Model} & \textbf{Baseline Fiedler} & \textbf{Semantic Fiedler} & \textbf{Hedges $g$} \\
\midrule
DistilGPT-2   & 0.76 & 0.79 & +0.34 \\
GPT-2         & 0.77 & 0.81 & +0.42 \\
GPT-2 Medium  & 0.78 & 0.83 & +0.56 \\
\bottomrule
\end{tabular}
\end{table}

\vspace{-1em}

\begin{table}[ht!]
\centering
\caption{Substitution hallucinations vs.\ baseline: entropy and smoothness index at the final layer. Effect sizes are moderate.}
\scriptsize
\label{tab:stats_substitution}
\begin{tabular}{lccc}
\toprule
\textbf{Model} & \textbf{Baseline Entropy} & \textbf{Substitution Entropy} & \textbf{Hedges $g$} \\
\midrule
DistilGPT-2   & 0.72 & 0.75 & +0.40 \\
GPT-2         & 0.71 & 0.74 & +0.47 \\
GPT-2 Medium  & 0.70 & 0.73 & +0.51 \\
\bottomrule
\end{tabular}
\end{table}

\subsubsection{Limitations.} While logical hallucinations clearly exceed baseline variability, semantic hallucinations often remain within factual variance for primary metrics (HFER, entropy, SMI). Their detection relies on subtler secondary signatures (Fiedler drift). This indicates that variance-based thresholds are insufficient: future work should develop adaptive, layerwise statistical detectors and account for multiple comparisons.

\subsection{Spectral Hallucination Detector}
To demonstrate practical utility, we implement a simple detector using normalized last-layer Fiedler z-scores:
\begin{equation}
\text{SHD}(x) = \mathbf{1}[z_{\text{fid}}(x) > \tau_d], \quad z_{\text{fid}}(x) = \frac{f_{\text{last}}(x) - \mu_{\text{fid}}}{\sigma_{\text{fid}}}
\end{equation}
where $f_{\text{last}}(x)$ is the final-layer Fiedler value, $\mu_{\text{fid}}, \sigma_{\text{fid}}$ are baseline statistics, and $\tau_d$ are domain-specific thresholds optimized per semantic domain. Table~\ref{tab:shd_performance} shows detection performance on 80 test samples, demonstrating that spectral signatures enable effective hallucination detection beyond theoretical analysis.

\vspace{-1em}
\begin{table}[ht!]
\centering
\caption{Hallucination detection performance on 80 test samples (50 factual, 30 hallucinations).}
\scriptsize
\label{tab:shd_performance}
\begin{tabular}{lcccc}
\toprule
& \textbf{SHD (domain)} & \textbf{Perplexity~\cite{lee2022factuality}} & \textbf{SelfCheckGPT-style~\cite{manakul2023selfcheckgpt}} \\
\midrule
\textbf{Accuracy} & \textbf{88.75\%} & 75.00\% & 65.00\% \\
\bottomrule
\end{tabular}
\end{table}

\vspace{-1em}\subsection{Interpretation}

Experiments reveal universal spectral convergence for factual reasoning (energy mountain, entropy dip, smoothness plateau, connectivity rise). Hallucinations, however, diverge: logical errors destabilize spectra, while semantic ones stay mostly stable but show entropy increase, smoothness loss, and Fiedler drift~\cite{Dale2024}. Newer models behave differently: Qwen2.5-7B, for instance, collapses late-layer connectivity, highlighting model-dependent spectral responses~\cite{Bai2023qwen}.

\begin{figure}[ht!]
\centering
\includegraphics[width=0.80\columnwidth]{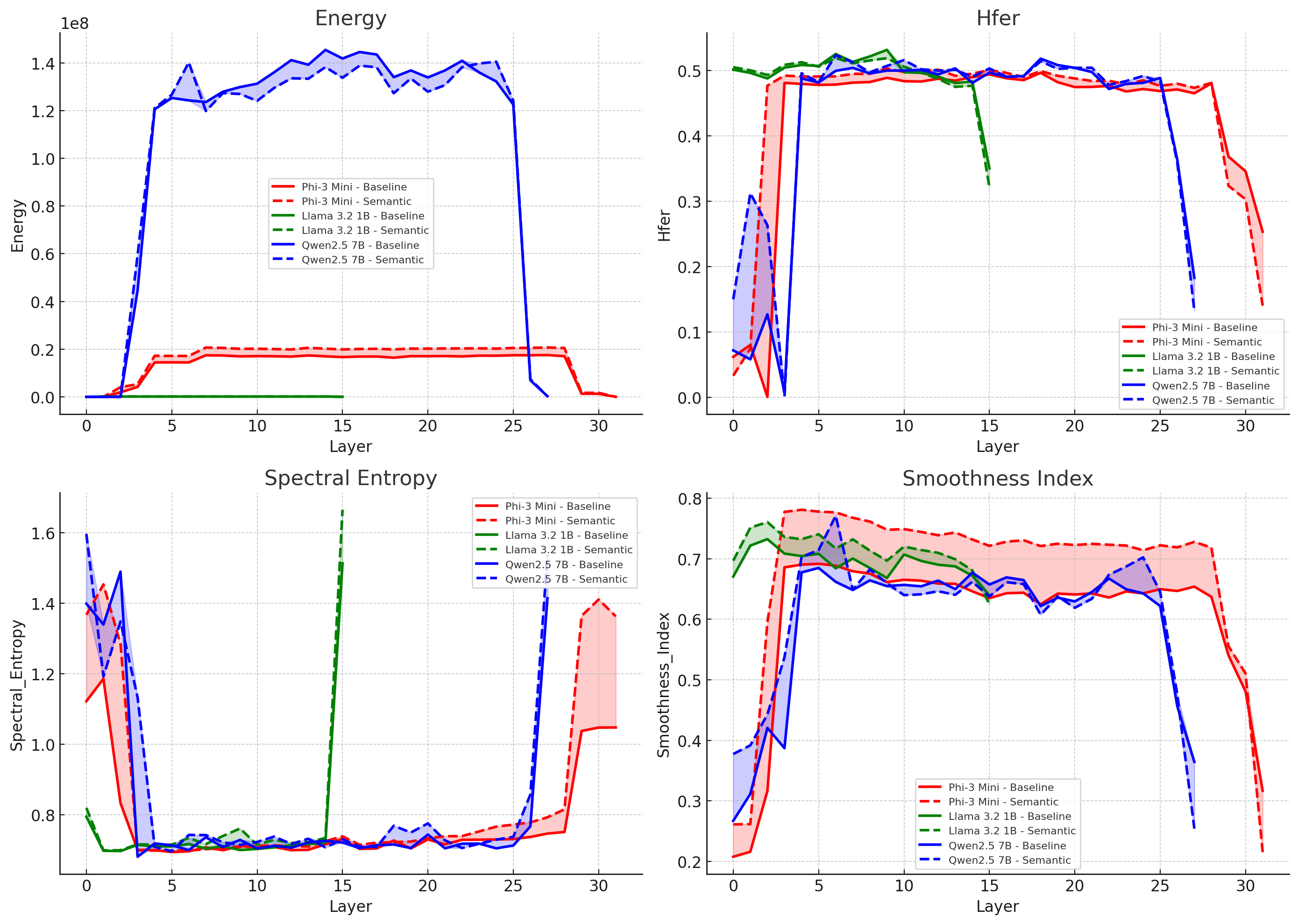}
\caption{Baseline vs.\ semantic hallucinations across new architectures (Phi-3 Mini, LLaMA-3.2 1B, Qwen2.5-7B). Error bands ($\pm$1 SD) are derived from factual runs. Semantic hallucinations remain within baseline variance for most metrics but diverge systematically in entropy, smoothness, and connectivity.}
\label{fig:new_models_baseline_semantic}
\end{figure}

% \begin{figure}[ht!]
% \centering
% \includegraphics[width=0.80\columnwidth]{figures/fiedler_baseline_vs_semantic_new_models.pdf.png}
% \caption{Fiedler trajectories for baseline vs.\ semantic hallucinations. While Phi-3 Mini and LLaMA-3.2 1B show modest drift, Qwen2.5-7B exhibits a sharp collapse, highlighting connectivity breakdown as a potential hallucination marker.}
% \label{fig:fiedler_baseline_semantic}
% \end{figure}

% \vspace{-1em}
% \begin{table}[ht!]
% \centering
% \caption{Final-layer HFER. Semantic hallucinations lower high-frequency dominance compared to baselines, though effects remain moderate.}
% \scriptsize
% \label{tab:stats_hfer}
% \begin{tabular}{lcccc}
% \toprule
% \textbf{Model} & \textbf{Baseline Mean} & \textbf{Semantic Mean} & \textbf{SD} & \textbf{Hedges $g$} \\
% \midrule
% phi-3-mini    & 0.25 & 0.14 & ±0.12 & -0.89 \\
% llama-3.2-1b  & 0.35 & 0.32 & ±0.05 & -0.68 \\
% qwen2.5-7b    & 0.18 & 0.13 & ±0.13 & -0.34 \\
% \bottomrule
% \end{tabular}
% \end{table}

\vspace{-1em}

\begin{table}[ht!]
\centering
\caption{Final-layer spectral entropy. Semantic hallucinations consistently raise entropy, indicating greater disorder in token graphs.}
\scriptsize
\label{tab:stats_spectral_entropy}
\begin{tabular}{lcccc}
\toprule
\textbf{Model} & \textbf{Baseline Mean} & \textbf{Semantic Mean} & \textbf{SD} & \textbf{Hedges $g$} \\
\midrule
phi-3-mini    & 1.05 & 1.36 & ±0.25 & +1.55 \\
llama-3.2-1b  & 1.51 & 1.67 & ±0.23 & +0.72 \\
qwen2.5-7b    & 1.41 & 1.54 & ±0.25 & +0.49 \\
\bottomrule
\end{tabular}
\end{table}

% \begin{table}[ht!]
% \centering
% \caption{Final-layer smoothness index. Semantic hallucinations reduce smoothness, suggesting weakened structural regularity.}
% \scriptsize
% \label{tab:stats_smoothness_index}
% \begin{tabular}{lcccc}
% \toprule
% \textbf{Model} & \textbf{Baseline Mean} & \textbf{Semantic Mean} & \textbf{SD} & \textbf{Hedges $g$} \\
% \midrule
% phi-3-mini    & 0.32 & 0.22 & ±0.15 & -0.71 \\
% llama-3.2-1b  & 0.64 & 0.62 & ±0.03 & -0.55 \\
% qwen2.5-7b    & 0.36 & 0.25 & ±0.11 & -0.93 \\
% \bottomrule
% \end{tabular}
% \end{table}

\begin{table}[ht!]
\centering
\caption{Final-layer Fiedler values. Connectivity drift emerges as the most discriminative marker of semantic hallucinations, with Qwen2.5-7B showing a collapse far beyond baseline variance.}
\scriptsize
\label{tab:stats_fiedler}
\begin{tabular}{lcccc}
\toprule
\textbf{Model} & \textbf{Baseline Mean} & \textbf{Semantic Mean} & \textbf{SD} & \textbf{Hedges $g$} \\
\midrule
phi-3-mini    & 0.66 & 0.63 & ±0.09 & -0.21 \\
llama-3.2-1b  & 0.76 & 0.73 & ±0.07 & -0.43 \\
qwen2.5-7b    & 0.80 & 0.20 & ±0.31 & -2.35 \\
\bottomrule
\end{tabular}
\end{table}

% =========================================================
%            6. COMPUTATIONAL COMPLEXITY
% =========================================================
\section{Computational Complexity}
Energy and smoothness require sparse matrix–matrix products $\mathcal{O}(\mathrm{nnz}(W)\,d)$ per layer. Spectral entropy and $\HFER$ need partial spectral information; randomized Lanczos scales near-linearly in $\mathrm{nnz}(W)$ for a small number of eigenpairs~\cite{Halko2011}. For sequences up to 512 tokens, analysis completes in 10-60 seconds on standard GPUs, making the framework practical for real-time diagnostics.

% =========================================================
%            7. DISCUSSION AND FUTURE DIRECTIONS
% =========================================================

\vspace{-1em}\section{Discussion and Future Directions}

This work establishes spectral analysis as a principled tool for transformer interpretability~\cite{Belinkov2022}. The universal ``energy mountain'' highlights consistent mechanisms of reliable generation, while distinct spectral fingerprints of errors enable diagnostic use.

Future directions include extending analysis to building adaptive detectors for real-time monitoring~\cite{Manakul2023}, and studying larger architectures. While the present study focuses on classification-style tasks, preliminary evidence suggests that linguistic structure may also shape spectral trajectories, pointing to connections between spectral geometry and human-interpretable constructs. This establishes spectral analysis as both theoretically grounded and practically useful for LLM understanding~\cite{Huang2023}.

% =========================================================
%                       8. CONCLUSION
% =========================================================
\vspace{-1em}\section{Conclusion}

In summary, we presented a spectral graph processing framework that reveals both universal convergence patterns in factual reasoning and distinct spectral fingerprints of hallucinations. Beyond theoretical insight, we showed that spectral markers enable a practical hallucination detector that outperforms strong baselines. Logical hallucinations destabilize spectra, semantic hallucinations manifest as connectivity drift and entropy rise, and substitution errors exhibit intermediate perturbations. Together, these findings establish spectral geometry as both an interpretive lens and a diagnostic tool for monitoring large language models.

% =========================================================
%                       REFERENCES
% =========================================================
% \bibliographystyle{ieeetr}
% \bibliography{refs}

\end{document}